\documentclass[11pt]{article}
\usepackage{coling2020}
\usepackage{times}
\usepackage{url}
\usepackage{latexsym}
\usepackage{soul}
\usepackage{url}
\usepackage{graphicx}
\usepackage{amsmath}
\usepackage{amsthm}
\usepackage{booktabs}
\usepackage{algorithm}
\usepackage{algorithmic}
\usepackage{bm}
\usepackage{multirow}
\usepackage{float}
\usepackage{verbatim}
\usepackage{graphicx}
\usepackage{subfigure} 
\usepackage{amsmath}
\usepackage{amssymb}
\usepackage{amsthm}
\colingfinalcopy % Uncomment this line for the final submission

\title{Knowledge Graph Embedding with Atrous Convolution and  Residual Learning}

\author{Feiliang Ren, Juchen Li, Huihui Zhang, Shilei Liu, Bochao Li, Ruicheng Ming, Yujia Bai \\
  School of Computer Science and Engineering, Northeastern University \\
    Shenyang,   110169, China\\
  {\tt renfeiliang@cse.neu.edu.cn} \\
}

\date{}

\begin{document}
\maketitle
\begin{abstract}
Knowledge graph embedding is an important task and it will benefit lots of downstream applications. Currently, deep neural networks  based methods achieve state-of-the-art performance. However, most of these existing methods  are very  complex and need much time for training and inference. To address this issue, we propose a simple but effective atrous convolution based knowledge graph embedding method. Compared with  existing state-of-the-art methods, our method has following main characteristics. First, it effectively increases feature interactions by using atrous convolutions. Second, to address the \emph{original information forgotten} issue and vanishing/exploding gradient issue, it uses the residual learning method. Third, it has simpler structure but much higher parameter efficiency.  We evaluate our method on six benchmark datasets with different evaluation metrics. Extensive experiments show that our model is very effective. On these diverse datasets, it achieves better results than the compared state-of-the-art methods on most of evaluation metrics. The source codes of our model could be found at \url{https://github.com/neukg/AcrE}. % on these  datasets.
\end{abstract}

\section{Introduction}
Knowledge graph is a kind of valuable knowledge bases and it is important for many AI-related applications. Generally, a KG stores factual knowledge in the form of structural triplets like $<$\emph{h, r, t}$>$, which means there is a kind of \emph{r} relation from \emph{h} (head entity) to \emph{t} (tail entity). Nowadays, great achievements have been made in building large scale KGs. Usually a KG may contain millions of entities and billions of relational facts. However, there are still two major difficulties that prohibit the availability of KGs. First, although most  existing KGs contain large amount of triplets, they are far from completeness. Second, most existing KGs are stored in symbolic and logical formations while applications often involve numerical computing in continuous spaces. To address these two issues, researchers proposed {\em knowledge graph embedding} (KGE) methods that aim to learn a kind of  embedding representations for a KG’s items (entities and relations) by projecting these items into some continuous low-dimensional spaces. Generally, different kinds of KGE methods mainly differ in how to view the role of relations in the projected spaces. For example, translation based methods (\emph{TransE} \cite{Bordes:2013}, \emph{TransH} \cite{Wang:2014}, \emph{TransR} \cite{Lin:2015a}, \emph{TransD} \cite{Ji:2015}, et al.) view the relation in a triplet as a translation operation from the head entity to the tail entity. Other KGE methods view relations as some kind of combination operators that link head entities and tail entities. For example, \emph{HolE}~\cite{Nickel:2016} employs a circular correlation function as the combination operator in the project space. \emph{ComplEx} \cite{Trouillon:2016} makes use of complex valued embeddings and takes the matrix decomposition as the combination operator. \emph{RT} \cite{Wang:2019}uses Tucker decomposition for KGE. \emph{RotateE} \cite{Sun:2019} use the rotation operation in the complex space as the combination operator. Experimental results show  these existing  methods have strong feasibility and robustness in solving the mentioned two issues.

Recently, deep  neural networks (DNN) based KGE methods~\cite{Dettmers:2018,Nguyen:2018,Yao:2020,Vashishth:2020a,Vashishth:2020b} push the performance of KGE to a soaring height. Compared with previous methods, this kind of methods can learn more effective embeddings mainly due to the powerful learning ability inherent in the DNN models. However, as pointed out by \newcite{Xu:2020} that existing research did not make a proper trade-off between the model complexity (the number of parameters) and the model expressiveness (the performance in capturing semantic information). Thus deep convolutional neural networks (DCNN) based methods  are achieving more and more research attention due to their simple but effective structure. 
However, \newcite{Chen:2018} point out that the DCNN based methods usually suffer from the reduced feature resolution issue that is caused by the repeated combination of max-pooling and down-sampling(“\emph{striding}”) performed at consecutive layers of {DCNNs}. This will result in feature maps with significantly reduced spatial resolution when {DCNN} is employed in a fully convolutional fashion.

To address this issue, we propose an atrous convolution  based KGE method which allows the model to effectively enlarge the field of view of filters almost without increasing the number of parameters or the amount of computations. To address the vanishing/exploding gradient issue inherent in the DNN based learning frame and the \emph{original information forgotten} issue when more convolutions used, we introduce residual learning in the our method.  We propose two learning structures  to integrate different kinds of convolutions  together: one is a  serial structure, and the other is a parallel  structure. 
We evaluate our method on six diverse benchmark datasets. Extensive experiments show that our method achieves better result than the compared state-of-the-art baselines under most evaluation metrics on these datasets. 

\section{Related Work}
Translation based KGE methods view the relation in a triplet as a translation operation from the head entity to the tail entity. These methods usually define a score function (or energy function) that has a form like $||$$\mathbf {h}$ + $\mathbf {r}$ - $\mathbf {t}$ $||$ to measure the plausibility of a triplet. During training, almost all of them minimize a margin based ranking loss function over the training data. \emph{TransE} \cite{Bordes:2013} is a seminal work in this branch. It directly takes the embedding space as a translation space. Formally, it tries to let $\mathbf {h}$ + $\mathbf {r}$  $\approx$   $\mathbf {t}$ if $<$\emph{h, r, t}$>$ holds. \emph{TransH} \cite{Wang:2014} models a relation as a hyperplane together with a translation operation on it. \emph{TransR} \cite{Lin:2015a} models entities and relations in distinct spaces, i.e., the entity space and multiple relation spaces. \emph{TransD} \cite{Ji:2015} models each entity or relation by two vectors. \emph{TranSparse} \cite{Ji:2016} mainly considers the heterogeneity property and the imbalance property in KGs.  \emph{PTransE} \cite{Lin:2015b} integrates relation paths into a TransE model. \emph{ITransF} \cite{Xie:2017} uses a sparse attention mechanism to discover hidden concepts of relations and to transfer knowledge through the sharing of concepts. Recently, researchers also employ the methods of combining different distance functions together for KGE. For example, \newcite{Sadeghi:2019} proposed a multi distance embedding (\emph{MDE}) model, which consists of several distances as objectives. 

Bilinear KGE models use different kinds of combination operators other than the translation. For example, \emph{HolE} \cite{Nickel:2016} employs a circular correlation as the combination operator. \emph{ComplEx} \cite{Trouillon:2016} makes use of complex valued embedding and takes matrix decomposition as the combination operator. Similar to \emph{ComplEx}, \emph{RotatE} \cite{Sun:2019} also use a complex space where each relation is defined as a rotation from the source entity to the target entity. \newcite{Xu:2019} proposed DihEdral for KG relation embedding. By leveraging the desired properties of dihedral group, their method could support many kinds of relations like symmetry, inversion, etc. \newcite{Wang:2019} propose the \emph{Relational Tucker3(RT)} decomposition for multi-relational link prediction in knowledge graphs. 

Other work, \emph{KG2E} \cite{He:2015} uses a density-based method to model the certainty of entities and relations in a space of multi-dimensional Gaussian distributions. \emph{TransG} \cite{Xiao:2016} mainly addresses the issue of multiple relation semantics.

%Text-aware methods use extra textual resources. In this line of methods, researchers mainly focus on employing textual descriptions of entities as a kind of supplementary information. Most of the text-aware methods consist of two basic components: one is to model triplet facts, and the other is to model textual descriptions. For the former one, the TransE model is often used. That is to say, the main difference among existing text-aware methods mainly lies in the latter one. For example, \emph{DKRL} \cite{Xie:2016a} uses convolutional neural networks (\emph{CNN}) to model entities’ textual description information. \emph{SSP} \cite{Xiao:2017} uses a topic model to encode entities’ textual description information. \newcite{Xu:2017} introduces three neural models to encode entities’ textual description information. The best performance is obtained when using the long short term memory network (\emph{LSTM}). \emph{TKRL} \cite{Xie:2016b} use entities’ types as complementary information.

Recently, researchers begin to explore the DNN based methods for KGE and achieve state-of-the-art results. For example, \emph{ConvE} \cite{Dettmers:2018} uses 2D convolution over embeddings and multiple layers of nonlinear features to model KGs. \emph{ConvKB} \cite{Nguyen:2018} also use convolutional neural network for KGE. \emph{ConMask} \cite{Shi:2018}uses relationship-dependent content masking, fully convolutional neural networks, and semantic averaging to extract relationship-dependent embeddings from the textual features of entities and relations in KGs.  More recently, 
\newcite{Guo:2019} studied the path-level KG embedding learning and proposed recurrent skipping networks(\emph{RSNs}) to remedy the problems of using sequence models to learn relational paths.  \newcite{Yao:2020} integrate BERT \cite{Devlin:2019} into the KGE model. \newcite{Wang:2020} propose CoKE which uses Transformer \cite{Vaswani:2017}. \newcite{Vashishth:2020a} extend {\em ConvE} by increasing the interactions with feature permutation, feature reshaping, and circular convolution.

Most recently, graph based neural network(\emph{GNN}) methods are achieving more and more attentions. ~\newcite{Schlichtkrull:2018} propose {\em R-GCN}, which is a graph based DNN model that uses neighboring information of each entity.   ~\newcite{Bansal:2019} propose \emph{A2N}, an attention-based model  based on graph neighborhood. ~\newcite{Shang:2019} propose a weighted graph convolutional network based method that mainly utilizes learnable relational specific scalar weights during GCN aggregation. ~\newcite{Ye:2019} propose \emph{VR-GCN}, which is an extention of graph convolutional networks for embedding both nodes and relations. ~\newcite{Shang:2019} propose \emph{SACN} that takes the benefit of both GCN and \emph{ConvE}. ~\newcite{Vashishth:2020b}  propose \emph{CompGCN} which jointly embeds both nodes and relations in a relational graph.

However, as pointed out by \newcite{Xu:2020} that most of these existing DNN-based or GNN-based KGE methods are very complex and time-consuming, which prevents them be used in some on-line or real-time application scenarios.

\begin{figure}[t]
	\centering\subfigure[]{\includegraphics[width=0.9\textwidth]{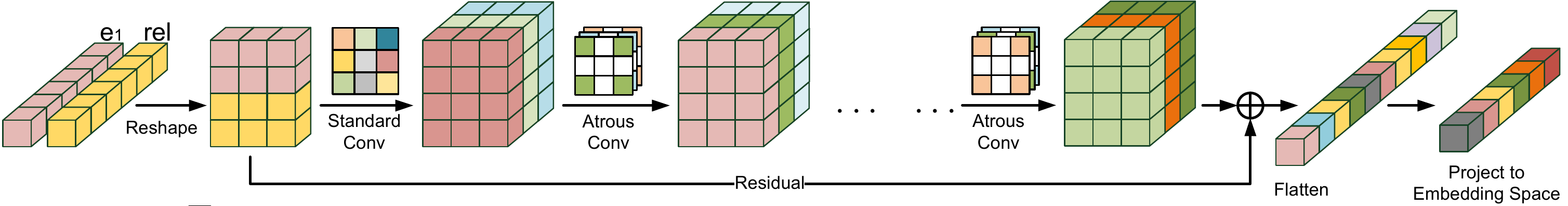}}
	\centering\subfigure[]{\includegraphics[width=0.9\textwidth]{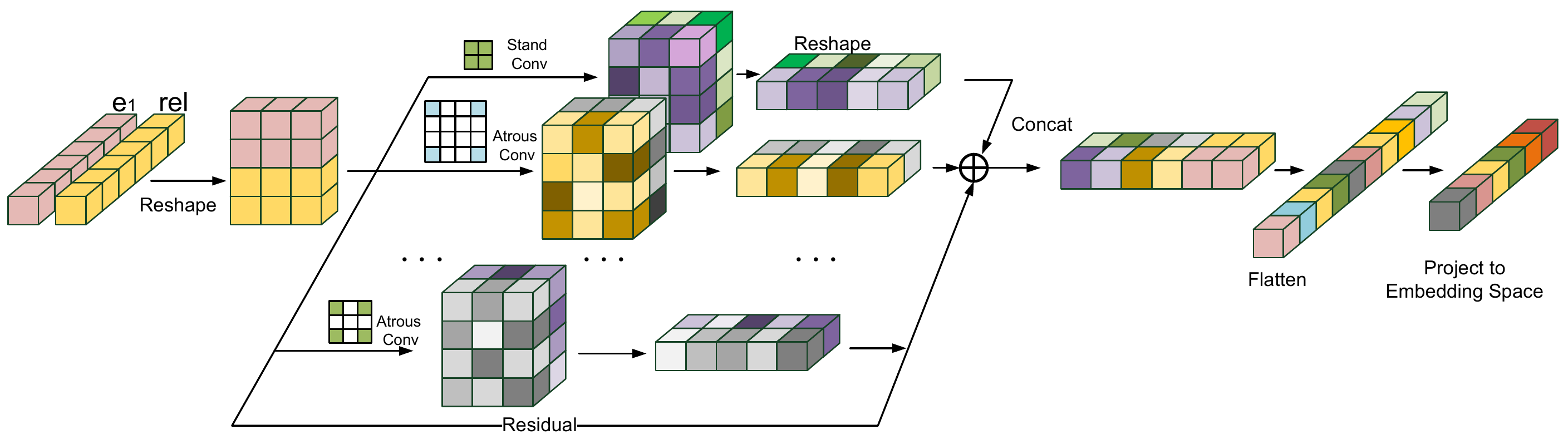}}
%	\subfigure[]{\includegraphics[width=0.45\textwidth]{2.pdf}} 
%	\qquad \qquad
%	\qquad
%	\subfigure[]{\includegraphics[width=0.45\textwidth]{3.pdf}}

\begin{comment}	
	\qquad
	\subfigure[]{\includegraphics[width=0.5\textwidth]{1.pdf}}
	\qquad
	\raisebox{0cm}
	{\hsize=0em
		\vbox{% 
			\subfigure[]{\includegraphics[width=0.44\textwidth]{2.pdf}} 
			\qquad \qquad
			
			\qquad
			\subfigure[]{\includegraphics[width=0.44\textwidth]{3.pdf}}
		}
	}
\end{comment}
	\caption{(a): Structure of  {\em Serial AcrE}; (b): Structure of  {\em Parallel AcrE}.}%The experimental results for $R_t= \exp \left( \left[1, -0.5, -2\right]^T \right)$.
	\label{fig:frame}
%}

\end{figure}

\section{AcrE Model}
We denote our model as \emph{AcrE} (the abbreviation of \textit{\textbf{A}}trous \textit{\textbf{C}}onvolution and \textit{\textbf{R}}esidual \textit{\textbf{E}}mbedding). In this study, we design  two structures to integrate the standard convolution and atrous convolutions together. One is a serial structure as shown in Figure \ref{fig:frame} (a), and the other is a parallel structure as shown in Figure \ref{fig:frame} (b). We will introduce them one by one in the following.  
%There are two main components in our AcrE model. One (the upper dashed red rectangle box part in Figure \ref{fig:frame} (a)) is a self-contained atrous convolution based KGE model that can be called as the basic \emph{AcrE} model; the other (the bottom dashed blue rectangle box part in Figure \ref{fig:frame} (a)) is a boosted \emph{AcrE} model that integrates entities’ multi-source textual descriptions.

\subsection{Serial AcrE Model}
In the \emph{Serial AcrE} model, the standard convolution and  atrous convolutions are organized in a serial manner. As shown in Figure \ref{fig:frame} (a),  the output of one convolution will be taken as the input of its subsequent adjoining  convolution. In this model, the embeddings of an entity and its relation are first reshaped into a 2-Dimension representation, then a standard convolution and several atrous convolutions are performed serially.  Next, the outputted embeddings  of the last atrous convolution and the initial embeddings are combined by a residual learning based method. The combined  embeddings are flattened into a vector. This vector  is then used as features to get the probability distributions for the entity candidates.

\noindent\textbf{2D Embedding Representation} For a triplet $<$\emph{h, r, t}$>$, we denote $\mathbf {h}$, $\mathbf {r}$ and $\mathbf {t}$ as their corresponding embedding representations. \emph{ConvE}  points out that a 2-Dimension (\emph{2D} for short) convolution operation is better than a \emph{1D}-convolution operation because a \emph{2D}-convolution increases the expressiveness of a CNN model through additional points of interaction between embeddings. Thus follow \emph{ConvE}, we also use a \emph{2D} convolution in our model. To this end, the embedding concatenation of an entity  and its linked relation  is reshaped into a \emph{2D} embedding representation. 

Specifically, we use  $\tau$ to denote a \emph{2D} reshaping function and use $\mathbf{e}$ to denote the embedding of an entity \emph{e}. If $\mathbf {e, r} \in \mathbb{R}^{m}$,  $\tau(\mathbf {[e;r]})\in \mathbb{R}^{n_1 \times n_2}$  where $2 \times m = n_1 \times n_2$. In this study, we use [$\mathbf {e;r}$] to denote the concatenation of $\mathbf {e}$  and $\mathbf {r}$. 

\noindent\textbf{Standard Convolution based Learning} After the 2D reshaping process, a standard convolution operation is performed with Equation \ref{eq:scon}.
\begin{equation}
\mathbf{C}^i_0 = \mathbf{\omega}^i_0\star \tau(\mathbf {[e;r]}) + \mathbf{b}^i_0  \label{eq:scon}
\end{equation}

where $\star$ denotes convolution operation, $\mathbf{\omega}^i_0 \in \mathbb{R}^{k \times k}$ is the $i$-th filter and $\mathbf b_0^i$ is the $i$-th bias vector.

Then the outputs of these filters are stacked to form the output of the standard convolution learning. We denote the final output of this standard convolution learning as $\mathbf{C_0}$ , which could be simply written as $\mathbf{C}_0 = \left[\mathbf{C}^1_0 : \mathbf{C}^2_0: \mathbf{C}^3_0: ... :\mathbf{C}^F_0\right]$  and $F$ is the number of filters used.

% we use $\left[\mathbf{C}^0_0 ; \mathbf{C}^1_1; \mathbf{C}^2_1; ... \right]$ to denote the stack of  each filter's outputs.

%$\mathbf{C}_0 = \left[\mathbf{C}^0_0 ; \mathbf{C}^1_1; \mathbf{C}^2_1; ... \right]$ constitutes the output of the standard convolution layer.

%After this standard convolution operation, a vector reshaping function is used to reshape $\bm C$ to a vector. Then a linear transformation is performed to generate a new representation (denoted as $\bm c$) for the input. This process is defined with Equation \ref{eq:vec}.
%\begin{equation}
%\bm{c} = vec(\bm{C}) * \bm{W} \label{eq:vec}
%\end{equation}

%where $vec(\bm{C})$ is the vector reshaping function that converts $\bm C$ to a vector, and $\bm W$ is a transformation matrix. In this study, we use the same reshaping method as used in \emph{ConvE}.

It should be noted that we don’t perform a max-pooling operation that is often used in traditional CNN models. This is because the input of our model is always an entity and a relation. Thus the length of the convolution output is fixed. It is unnecessary to use a max-pooling to generate a new length-fixed representation. Our in-house experiments show that there is no obvious performance difference  between with and without a max-pooling operation. %most of the time. But sometimes there is slight performance degradation when the max-pooling is used.

\noindent\textbf{Atrous Convolution based Learning} %In this module, an atrous convolution based method is used to learn a new embedding representation for an entity and its linked relation. 
Atrous convolution, also called as dilated convolution, inserts some holes (zeros) in the input during convolution. Given an input vector $\mathbf {x}$  with a filter vector $\mathbf {w}$  of length \emph{K}, the output vector $\mathbf {y}$  of an atrous convolution is computed with Equation \ref{eq:atr}.

\begin{equation}
y_i = \sum_{k=1}^{K}x_{i+l \times k}\times w_k \label{eq:atr}
\end{equation}

Here \emph{l} (the atrous rate parameter) means the stride with which we sample the input. Obviously, the standard convolution is a special case of the atrous convolution when \emph{l} is set to 1. 

Specifically, in the \emph{Serial AcrE} model, an atrous convolution takes the output of its previous convolution as input, and output a new result with Equation \ref{eq:atrf}.

\begin{equation}
\mathbf{C_t} = \mathbf{\omega_t}\star {\mathbf C_{t-1}} + \mathbf{b_t}  \label{eq:atrf}
\end{equation}

where $\mathbf{ C_{t-1}}$ is the output of previous convolution operation,  $\mathbf{\omega_t}$ and $\mathbf{b_t}$ are the filter and bias vector respectively in the \emph{i-th} convolution. 

\noindent\textbf{Feature Vector Generation} In the \emph{Serial AcrE} model, different kinds of convolutions are performed one by one. Each convolution will extract some interaction features from the output of its previous convolution. Thus   the mined features would ``\emph{forget}" more and more original input information as  convolutions performed. However, the original information is the foundation of all mined features, so ``\emph{forget}" them will increase the risk that the mined features are actually irrelevant to what are needed. We call this phenomenon as \emph{original information forgotten} issue. Besides, there is an inherent  vanishing/exploding gradient issue in the deep networks. Here  we use the residual learning method \cite{He:2016} to add original input information back so as to address both issues. Then the result of residual learning is  flattened into a feature vector. Specifically, the whole process is defined with Equation \ref{eq:relu}.
\begin{equation}
\mathbf{o} = {Flatten}(\text{ReLU}(\mathbf{C}_T+\tau(\mathbf {[e;r]}))) \label{eq:relu}
\end{equation}

where $\mathbf{C}_T$ is the output of last atrous convolution, and $T$ is the number of atrous convolutions.

%Here “\emph{AtrousConvBlock}” means the function that maps an input by a standard convolution layer and several atrous convolution layers.

%Finally, different convolution operations’ outputs need to be combined into ONE unified representation. There are two widely used methods for this. One is a gate method; the other is a simple concatenation operation. We compared these two methods in experiments and found there is almost no difference in performance. So we use the simpler concatenation method here. The generated new representation is also denoted as $\bm c$.

\noindent\textbf{Score Function} With the generated feature vector $\mathbf{o}$, we define the following  function to compute a score to measure the degree of an entity candidate $t$ can form a correct triplet with the input $<$\emph{h,r}$>$. %, which means the vector $\bm c$ is matched with a candidate entity $\bm t$ via an inner product operation (denoted as $\odot$).
\begin{equation}
\psi(h,r,t) = \left(\mathbf{o}\mathbf{W}+\mathbf{b}\right)\mathbf{t}^\top \label{eq:score1}
\end{equation} 
%ψ(<h,r,t>) = c⊙t                           (5)

where $\mathbf{W}$ is a transformation matrix and $\mathbf{b}$ is a  bias vector. 
Then a sigmoid function is used to get the probability distribution over all candidate entities.
\begin{equation}
p\left(t|h,r\right) = sigmoid(\psi(h,r,t)) \label{eq:prob1}
\end{equation} 
%probt = softmax(ψ(<h,r,t>))                   (6)

\subsection{Parallel AcrE Model}
In the \emph{Parallel AcrE} model, the standard convolution and  atrous convolutions are organized in a parallel manner. As shown in Figure \ref{fig:frame} (b),  different kinds of convolutions are performed simultaneously, then their results are combined and flattened into a vector. Similar to the \emph{Serial AcrE} model, this vector  is  used as features to get the probability distributions for the entity candidates.

Compared with the \emph{Serial AcrE} model, most of the components in the \emph{Parallel AcrE} model have the same definitions except for the results integration and feature vector generation. We will introduce these two differences in the following part. 

\noindent\textbf{Results Integration} Different from the \emph{serial} structure, there will be multi results generated by different convolution operations. Accordingly, we need to integrate these results together. This process can be defined with following Equation \ref{eq:con}. 

\begin{equation}
\mathbf{C} = \mathbf{C_0} \oplus \mathbf{C_1} \oplus ...\oplus \mathbf{C_T} \label{eq:con}
\end{equation} 

where $\mathbf{C_0}$ is the result of standard convolution and $\mathbf{C_i}$ is the result of the \emph{i-th} atrous convolution, and $\oplus$ means a result integration operation. There are different kinds of integration methods. In this study, we explore two widely used  methods for this. One is an element-add operation based method,  the other is a  concatenation operation based method.  

\noindent\textbf{Feature Vector Generation} As shown in Figure \ref{fig:frame}, the final output of the whole convolution learning is followed by  a transformation operation. Then the results are flattened into the feature vector. Specially, the process can be written with Equation \ref{eq:relu1}, where $\mathbf{W_1}$ is the transformation matrix.
\begin{equation}
%\bm{c}  = vec(Relu(AtrousConvBlock([\bm{[e;r]})+[\bm {e;r}])) \label{eq:relu}
\mathbf{c}  = Flatten(\mathbf{W_1}Relu(\mathbf{C}+\tau(\mathbf{[e;r]}))) \label{eq:relu1}
\end{equation}

\subsection{Training}
Different from other KGE methods that often use a max-margin loss function for training, most neural networks based KGE methods (like \emph{ProjE}, \emph{ConvE}, etc.) often use the following two kinds of ranking loss functions. One is a kind of binary cross-entropy loss that the ranking scores are calculated independently (\emph{pointwise} ranking method), and the other is a kind of softmax regression loss that considers the ranking scores collectively (\emph{listwise} ranking method). Both \emph{ProjE} and \emph{ConvE} show that the latter one achieves better experimental results. In \emph{AcrE}, we define a same \emph{listwise} loss function as used in \emph{ConvE}.
\begin{equation}
\mathcal{L}= -\frac{1}{N} \sum_{i=1}^N \left[{t}_i \log p\left({t}_i|h,r\right) +(1-{t}_i)\log \left(1 - p\left({t}_i|h,r\right)\right) \right] \label{eq:loss}
\end{equation} 

%Loss(prob,t)=-1/N∑i(tilog(probi)+(1-ti)log(1-probi))(17)
where $\mathbf{t}$ is a label vector whose elements are ones for relationships that exist and zero otherwise, and \emph{N} is the number of entities in a KG. This loss function takes one \emph{(h,r)} pair and scores it against all entities simultaneously. Thus our model is very fast for both training and inference.

%During training, the embeddings of both entities and relations are initialized with the results of \emph{TransE} (this  method is also used by other KGE methods like \emph{TransR}, \emph{TransD}, \emph{TranSparse}, \emph{DKRL}, \emph{TKRL}, etc.). Other parameters, including transformation matrixes, bias terms, and gate vectors, are randomly initialized. All the parameters are tuned using a back propagation method. Adam is used as an optimizer.%\emph{ConvE} pointed out that this loss function takes one \emph{(h,r)} pair and scores it against all entities simultaneously

\section{Experiments and Analyses}

\subsection{Experiment Settings}
{\bf Datasets} We evaluate our method on six widely used benchmark datasets. The first two are WN18 \cite{Bordes:2014} and FB15k \cite{Bordes:2014}. The second two are  WN18RR and FB15k-237 \cite{Dettmers:2018}, which are two variant datasets for WN18 and FB15k to  avoid test leakage. The rest two are  Alyawarra Kinship~\cite{Lin:2018} and DB100K \cite{Ding:2018}, both are new datasets proposed in recent years. Some statistics of these six datasets are shown in Table \ref{tab:stas}.

%We evaluate the effect of multi-source descriptions on FB15k and FB15k-237. \emph{DBpedia} and \emph{Wikidata} are used as the data sources for downloading the needed descriptions. Different from Freebase whose entities are mainly named entities, the entities in WordNet are mainly common words. Thus few researchers evaluate the effective of descriptions on WordNet for there is no proper descriptions for  it. 

\noindent\textbf{Evaluation Task} We use \emph{link prediction}, one of the most frequently used benchmark evaluation tasks for KGE methods, to evaluate our model. \emph{Link prediction} is to predict the missing \emph{h} or \emph{t} for a correct triplet $<$\emph{h, r, t}$>$, i.e., predict \emph{t} given $<$\emph{h, r}$>$ or predict \emph{h} given $<$\emph{r, t}$>$. Rather than requiring one best answer, this task emphasizes more on ranking a set of candidate entities from the KG. \emph{Hits@k} and \emph{MRR} are often used as the evaluation metrics. 

In experiments, %the embeddings are initialized with the results of \emph{TransE} (this  practice is widely used by existing methods like \emph{TransR}, \emph{TransD}, \emph{TransH}, etc.).
all the parameters, including initial embeddings, transformation matrices, and bias vectors,  are randomly initialized. Hyper-parameters are selected by a grid search on the validation set. All the results are reported when 3 atrous convolutions used for both learning structures. 
% are tuned using a back propagation method. Adam is used as an optimizer. All the 

%In experiments, we follow the same protocol used in other methods (like \emph{TransE}, \emph{TransH}, etc): for each testing triplet $<$\emph{h, r, t}$>$, we replace the tail \emph{t} by every entity \emph{e} in the KG and calculate a score (according to the function $\psi$, Equation \ref{eq:score1}) on the corrupted triplet $<$\emph{h, r, e’}$>$. Ranking the scores in ascending order, we then get the rank of the original correct triplet. Similarly, we can get another rank for $<$\emph{e’, r, t}$>$ by corrupting the head \emph{h}. We call this evaluation setting as “\emph{Raw}”. However, if a corrupted triplet exists in the KG, it should be regarded as a correct triplet and ranking it before the original triplet is not wrong. To eliminate this factor, previous work (\emph{TransH/D/R}, etc.) often removes those corrupted triplets that exist in either training, validation, or test set before getting the rank of each test triplet. Such setting is called “\emph{Filt}”. Obviously, this setting can reflect the true capability of a KGE method, thus almost all of the state-of-the-art methods report their experimental results under the “\emph{Filt}” setting. In this work, we also report our experimental results under the “\emph{Filt}” setting.

%\begin{comment}

\begin{table}[t]
	\centering
	%	\caption{Add caption}
	%	\small
	%	\makebox[\linewidth]
	\begin{tabular}{llllll}
		\toprule
		\multirow{2}[4]{*}{Dataset} & \multicolumn{1}{c}{\multirow{2}[4]{*}{\#R}} & \multicolumn{1}{c}{\multirow{2}[4]{*}{\#E}} & \multicolumn{3}{c}{\#Triplet} \\
		\cmidrule{4-6}          &       &       & \multicolumn{1}{l}{Train} & \multicolumn{1}{l}{Valid} & \multicolumn{1}{l}{Test} \\
		\midrule
		DB100K  & 470    & 99,604 & 597,572 & 50,000 & 50,000 \\
		WN18  & 18    & 40,943 & 141,442 & 5,000 & 5,000 \\
		FB15k & 1,345 & 14,951 & 483,142 & 50,000 & 59,071 \\
		WN18RR & 11    & 40,943 & 86,835 & 3,034 & 3,134 \\
		FB15k-237 & 237   & 14,541 & 272,115 & 17,535 & 20,446 \\
		Kinship & 25    & 104   & 8,544 & 1,068 & 1,074 \\
		\hline
	\end{tabular}%
	\caption{Dataset statistics.}
	\label{tab:stas}%
\end{table}%
\begin{table}[t]
	\centering
	
	\begin{tabular}{lllll}
		\toprule
		Model & \multicolumn{1}{l}{MRR} & \multicolumn{1}{l}{H@1} & \multicolumn{1}{l}{H@3} & \multicolumn{1}{l}{H@10} \\
		\midrule
		TransE\cite{Bordes:2013} & 0.111 & 1.6   & 16.4  & 27 \\
		DistMult\cite{Yang:2015} & 0.233 & 11.5  & 30.1  & 44.8 \\
		HolE\cite{Nickel:2016}  & 0.26  & 18.2  & 30.9  & 41.1 \\
		ComplEx\cite{Trouillon:2016} & 0.242 & 12.6  & 31.2  & 44 \\
		Analogy\cite{Liu:2017} & 0.252 & 14.2  & 32.3  & 42.7 \\
		RUGE\cite{Guo:2018}  & 0.246 & 12.9  & 32.5  & 43.3 \\
		ComplEx-NNE+AER\cite{Ding:2018} & 0.306 & 24.4  & 33.4  & 41.8 \\
		Sys-SEEK\cite{Xu:2020} & 0.306 & 22.5  & 34.3  & 46.2 \\
		SEEK\cite{Xu:2020}  & 0.338 & 26.8  & 37    & 46.7 \\
		\midrule
		
		{\bf AcrE (Serial)} & {\bf 0.399} & {\bf 30.4}  & {\bf 45.3}  & {\bf 57.0} \\
		{\bf AcrE (Parallel)} & {\bf 0.413} & {\bf 31.4}  & {\bf 47.2}    & {\bf 58.8} \\
		\hline
	\end{tabular}%
	\caption{Experimental results on DB100k. All the compared results are taken from \newcite{Xu:2020}.}
	\label{tab:main0}%
\end{table}%

\begin{table}[thbp]
	%	\centering
	%	\small
	%	\makebox[\linewidth]
	\resizebox{\textwidth}{!}
	{\begin{tabular}{lllllllll}
			\hline
			\multicolumn{1}{r}{} & \multicolumn{4}{c}{FB15K}  & \multicolumn{4}{c}{WN18} \\
			\multicolumn{1}{r}{}  & \multicolumn{1}{l}{MRR} & \multicolumn{1}{l}{H@1} & \multicolumn{1}{l}{H@3} & \multicolumn{1}{l}{H@10} & \multicolumn{1}{l}{MRR} & \multicolumn{1}{l}{H@1} & \multicolumn{1}{l}{H@3} & \multicolumn{1}{l}{H@10} \\
			
			\hline

			TransE\footnotesize{\cite{Bordes:2013}}& {0.463} & {29.7} & {57.8} & {74.9} & {0.495} & {11.3} &{88.8} &{94.3}\\
			HolE\footnotesize{\cite{Nickel:2016}} & {0.524} & {40.2} & {61.3} & {73.9} & {93.8} & {93.0} &{94.5} &{94.9}\\
			ComplEx\footnotesize{\cite{Trouillon:2016}} & 0.692 & 59.9 & 75.9 &84.0   &0.941 &93.6 &94.5 &94.7      \\
			SimplE\footnotesize{\cite{Kazemi:2018}}  &0.727 & 66.0 & 77.3 &83.8   &0.942 &93.9 &94.4 &94.7     \\
			D4-Gumbel\footnotesize{\cite{Xu:2019}}  & 0.728 & 64.8 & 78.2 &86.4   &0.728 &64.8 &78.2 &86.4     \\
			D4-STE\footnotesize{\cite{Xu:2019}} & 0.733 & 64.1 & 80.3 &87.7        &0.733 &64.1 &80.3 &87.7 \\
			ConvE\footnotesize{\cite{Dettmers:2018}} & 0.657 & 55.8 & 72.3 &83.1    &0.942 &93.5 &94.7 &95.5     \\
			
			R-GCN\footnotesize{\cite{Schlichtkrull:2018}}  & 0.696 & 60.1 & 76.0 &84.2   &0.696 &60.1 &76.0 &84.2      \\
			%			TuckER\footnotesize{\cite{Balazevic:2019a}}  & 0.727 & 66.0 & 77.3 &83.8        &{\bf 0.953} &{\bf 94.9} &{\bf 95.5} &95.8 \\
			RotatE\footnotesize{\cite{Sun:2019}}  & 0.797 & 74.6 & 83.0 &88.4   &0.949 &94.4 &{\bf 95.2} &{\bf 95.9 }     \\
			
			RSNs\footnotesize{\cite{Guo:2019}} & 0.78 & 72.2 & - &87.3  &0.94 &92.2 &- &95.3      \\
			\midrule
			{\bf AcrE(Serial)} & {0.791} & {72.7} & {\bf 83.8} &{\bf 89.6}   & {\bf 0.950} &{\bf 94.6} &{\bf 95.3} &{\bf 95.9}     \\
			{\bf AcrE(Parallel)} & {\bf 0.815} & {\bf 76.4} & {\bf 85.2} &{\bf 89.8}   & {0.948} &{94.3} &{\bf 95.2} &{95.7}     \\
			\midrule
			\multicolumn{1}{r}{} & \multicolumn{4}{c}{FB15K-237}  & \multicolumn{4}{c}{WN18RR} \\
			\cmidrule{2-9}     
			\multicolumn{1}{r}{} &  \multicolumn{1}{l}{MRR} & \multicolumn{1}{l}{H@1} & \multicolumn{1}{l}{H@3} & \multicolumn{1}{l}{H@10} & \multicolumn{1}{l}{MRR} & \multicolumn{1}{l}{H@1} & \multicolumn{1}{l}{H@3} & \multicolumn{1}{l}{H@10} \\
			
			\midrule
			
			ConvE\footnotesize{\cite{Dettmers:2018}}     & 0.312  & 22.5  & 34.1  & 49.7    & 0.43 & 40  & 44    & 52  \\
			ConvKB\footnotesize{\cite{Nguyen:2018}}    & 0.243  & 15.5  & 37.1  & 42.1    & 0.249 & 0.057  & 41.7    & 52.4   \\
			R-GCN\footnotesize{\cite{Schlichtkrull:2018}}    & 0.164  & 10  & 18.1  & 30     & 0.123 & 8  & 13.7    & 20.7  \\
			%			TuckER\footnotesize{\cite{Balazevic:2019a}}    & 0.358  & {26.6}  & 39.4  & {\bf 54.4}    & 0.470 & 44.3  & 48.2   & 52.6  \\
			RotatE\footnotesize{\cite{Sun:2019}}     & 0.338  & 24.1  & 37.5  & 53.3   & 0.476 & 42.8  & {49.2}   & {\bf 57.1} \\
			D4-STE\footnotesize{\cite{Xu:2019}}   & 0.320  & 23.0  & 35.3  & 50.2    & {0.480} & {\bf 45.2}  & 49.1   & 53.6  \\
			
			SACN\footnotesize{\cite{Shang:2019}}     & 0.35  & 26.0  &{39.0}  &{54.0}     & 0.47 & 43.0  & 48.0   & 54.0  \\
			HypER\footnotesize{\cite{Balazevic:2019b}}    & 0.341  & 25.2  & 37.6  & 52.0   & 0.465 & 43.6  & 47.7   & 52.2  \\
			ConvR\footnotesize{\cite{Jiang:2019}}    & 0.350  & 26.1  & {38.5}  & 52.8    & 0.475 & {44.3} & 48.9   & 53.7  \\
			VR-GCN\footnotesize{\cite{Ye:2019}}   & 0.248  & 15.9  & 27.2  & 43.2   & - & -  & -   & -  \\
			RSNs\footnotesize{~\cite{Guo:2019}} & 0.280  & 20.2  & -  & 45.3   & - & -  & -   & -  \\
			DK-STE\footnotesize{~\cite{Xu:2019}}  & 0.320  & 23.0  & {35.3}  & 50.2   & {\bf 0.480} & {\bf 45.2}  & {49.1}   & {53.6}  \\
			%			InteractE\footnotesize{\cite{Vashishth:2020a}}    & {\bf 0.454}  & 26.3  & -  & 53.5    & 0.463 & 43.0  & -   & 52.8  \\
			CompGCN\footnotesize{\cite{Vashishth:2020b}}  & 0.355  & 26.4  & {39.0}  & 53.5   & 0.479 & {44.3}  & {\bf 49.4}   & {54.6}  \\

			%			DK-Gumbel\footnotesize{~\cite{Xu:2019}} & 0.300  & 20.4  & {33.2}  & 49.6   & 0.486 & {44.32}  & {\bf 50.5}   & {\bf 55.7}  \\
			%			MDE\footnotesize{~\cite{Sadeghi:2019}}  & 0.355  & 26.4  & {39.0}  & 53.5   & 0.479 & {44.3}  & {\bf 49.4}   & {\bf 54.6}  \\
			%			DRT\footnotesize{~\cite{Wang:2019}}  & 0.355  & 26.4  & {39.0}  & 53.5   & 0.479 & {44.3}  & {\bf 49.4}   & {\bf 54.6}  \\
			%			SRT\footnotesize{~\cite{Wang:2019}} & 0.355  & 26.4  & {39.0}  & 53.5   & 0.479 & {44.3}  & {\bf 49.4}   & {\bf 54.6}  \\			
			%			HAKE\footnotesize{\cite{Zhang:2020}}  & {-}   & 0.346  & 25.0  & 38.1  & 54.2 & -   & {\bf 0.497} & 45.2  & {\bf 51.6}   & {\bf 58.2}  \\
			%			CoKE\footnotesize{\cite{Wang:2020}}  & {448*}   & 0.364  & 27.2  & 40.0  & 54.9 & 4723*   & { 0.484} & 45.0  & { 49.6}   & { 55.3}  \\
			\midrule
			{\bf AcrE(Serial)} &  {0.352}  & {26.0}  & {38.8}  & {53.7}    & {0.463} & {42.9}  & {47.8}    & {53.4}   \\
			{\bf AcrE(Parallel)} &  {\bf 0.358}  & {\bf 26.6}  & {\bf 39.3}  & {\bf 54.5}    & {0.459} &{42.2}  & {47.3}    & {53.2}   \\
			\hline

	\end{tabular}}
	\centering
	{\begin{tabular}{lllll}
			\multicolumn{1}{r}{} & \multicolumn{4}{c}{Kinship}  \\
			\cmidrule{2-5} 
			\multicolumn{1}{r}{} &  \multicolumn{1}{l}{MRR} & \multicolumn{1}{l}{H@1} & \multicolumn{1}{l}{H@3} & \multicolumn{1}{l}{H@10}  \\
			
			\midrule
			ComplEx{\cite{Trouillon:2016}}     & 0.823  & 73.3  & 89.9  & 97.1     \\
			ConvE{\cite{Dettmers:2018}}     & 0.833  & 73.8  & 91.7  & 98.1     \\
			ConvKB{\cite{Nguyen:2018}}     & 0.614  &43.6 & 75.5  & 95.3    \\
			R-GCN{\cite{Schlichtkrull:2018}}     &0.109 &3 &8.8 &23.9    \\
			SimplE{\cite{Kazemi:2018}}     &0.752 &62.6 &85.4 &97.2    \\
			RotatE{\cite{Sun:2019}}     &0.843 &76.0 &91.9 &97.8    \\
			HAKE{\cite{Zhang:2020}}     &0.852 &76.9 &92.8 &98.0    \\
			InteractE{\cite{Vashishth:2020a}}     &0.777 &66.4 &87.0 &95.9    \\
			CompGCN{\cite{Vashishth:2020b}}     &0.778 &66.7 &86.8 &96.7    \\
			CoKE{\cite{Wang:2020}}     &0.793 &69.3 &87.8 &95.4    \\
			\midrule
			{\bf AcrE(Serial)} &  {\bf 0.864}  & {\bf 78.7}  & {\bf 93.1}  & {\bf 98.7}     \\
			{\bf AcrE(Parallel)} &  {\bf 0.864}  & {\bf 78.5}  & {\bf 93.9}  & {\bf 98.4}     \\
			%			{\bf AcrE(Parallel)} &  { 0.358}  & {\bf 26.6}  & {\bf 39.3}  & {\bf 54.5}     \\
			\hline

	\end{tabular}}
	\caption{Experimental results on the rest five benchmark datasets.}
	\label{tab:main}%
\end{table}

% Table generated by Excel2LaTeX from sheet 'Sheet9'

\subsection{Experimental Results}
\textbf{Overall Results} Table \ref{tab:main0} and \ref{tab:main}  show the experimental results on different datasets under different evaluation metrics. It should be noted that not all models report their results on all these six datasets, so  the compared  baselines on different datasets are different in these two tables. In subsequent part, all the experimental results for the  compared baselines  are taken from some latest published papers or their original papers. 
From these results we can draw following two conclusions. 

First,  our model is very robust and it significantly outperforms the compared state-of-the-art results under all  the evaluation metrics on all datasets  except for {WN18RR}. Especially on DB100K, FB15k, and Kinship, both \emph{AcrE (Serial)} and \emph{AcrE (Parallem)} outperform the compared baselines by a large margin under almost all the evaluation metrics.  As for WN18RR, our model still achieves very competitive results. Especially when compared with other DCNN-based KGE methods like \emph{ConvE} and \emph{ConvKB}, we can see that both the {\em Serial} and the \emph{Parallel} models perform much better.

Second, \emph{AcrE (Parallel)} performs  better than  \emph{AcrE (Serial)} in most  cases. We think this is mainly due to the reason that the \emph{Serial} structure based method suffers more from the \emph{original information forgotten} issue than the \emph{Parallel} structure based method.

%Third, mining expressive features is one of the key components for improving the performance of KGE methods.

%We think this is mainly because the atrous convolution increases features’ expressiveness through additional points of interaction by enlarging the field of view of filters. This is also in line with the motivations of \emph{ConvE} and \emph{InteractE}. %These experiments indicate that mining expressive features is one of the key components for improving the performance of KGE methods.

%\end{comment}
\noindent {\bf Detailed Results} 
We  conduct following two kinds of detailed experiments to further demonstrate the performance of our model. One is \emph{Head and Tail Prediction}, and the other is \emph{Prediction by Categories}.  

In the first kind of detailed experiments,  we  compare the performance of our model with several representative state-of-the-art baselines on {FB15k-237} for predicting missing head entities and predicting missing tail entities.  The results are summarized into Table \ref{tab:detail1}, from which we can see that our model outperforms the compared baselines again under all the evaluation metrics.
% We can also observe that there is a big performance gap between head prediction and tail prediction. For example, the performance gap between these two kinds of predictions is larger than about 20 in all the models under all evaluation metrics.  These results show that existing KGe models perform far better on forward prediction than backward prediction. This phenomenon is very confusing since both kinds of predictions have the same number of training samples. We think a possible reason is that human is more used to forward inference than backward inference. Thus they are more prone to design a model in line with this  habit.  In other word, the designed models are essentially biased to the forward prediction.

In the second kind of detailed experiments,  we  compare the performance of our model with several representative state-of-the-art baselines on {FB15k} for predicting by different categories.
%\footnote{ Researchers ofthen classify a KG’s relations into 4 types according to the type definitions given by TransH: For each relation $r$, compute averaged number of tails perhead ($tph_r$), and averaged number of head per tail ($hpt_r$ ). If $tph_r \le 1.5$ and $hpt_r \le 1.5$, $r$ is treated as 1-to-1. If $tph_r \geq 1.5$ and $hpt_r \geq 1.5$, $r$ is treated as m-to-n. If $hpt_r \le 1.5$ and $tph_r \geq 1.5$, $r$ is treated as 1-to-m. If $hpt_r \geq 1.5$ and $tph_r \le 1.5$, $r$ is treated as m-to-1.}. 
The results are shown in Table \ref{tab:detail2}. We can see that {\em ArcE}
% obtains more balanced performance among different kinds of relations: in {\em ArcE}, the performance gaps among different kinds of relations are less than that of in other methods. Especially, {\em ArcE} 
does much better than other compared baselines on almost all types of relations except the 1-to-1 relations. 
%These experimental results show that for a relation, the more head (tail) entities linked by it, the better \emph{ArcE} can do.
This merit is much important for real application scenarios where the complexer relations often take up large proportions. For example, in FB15k, one of the largest available KGs, the triplets of 1-to-1 are about 1.4\%, 1-to-n are about 8.9\%, n-to-1 are about 14.6\%, and m-to-n are about  75.1\%. %In other word, \emph{AcrE} does better for the triplets that take up the largest proportion. 
%From Table \ref{tab:detail2} we can also observe that when predicting the complex part of a relation, most of methods do much poor. For example, the Hits@10 scores for both the n-to-1 relations’ head prediction and the 1-to-n relations’ tail prediction are far lower than the scores of other kinds of prediction. This trend also exists in {\em ArcE}, but our method does much better than the compared baseline methods, which indicates that by increasing the interactions with atrous convolutions, our method can alleviate the data sparsity issue greatly.

%\noindent\textbf{Ablation Experiments} To demonstrate the contributions of different components in \emph{ArcE}, we conduct ablation experiments and the results are shown in Table \ref{tab:tab3}. 

\begin{table}[t]%[htbp]
	\centering
	\resizebox{\textwidth}{!}
	{\begin{tabular}{lp{0.5em}p{0.5em}p{0.5em}p{0.5em}p{0.5em}p{0.5em}p{0.5em}p{0.5em}}
			\hline
			\multicolumn{1}{r}{} & \multicolumn{4}{c}{Predict Head}  & \multicolumn{4}{c}{Predict Tail} \\
			\cmidrule{2-9} 
			\multicolumn{1}{r}{} &  \multicolumn{1}{l}{MRR} & \multicolumn{1}{l}{H@1} & \multicolumn{1}{l}{H@3} & \multicolumn{1}{l}{H@10}& \multicolumn{1}{l}{MRR} & \multicolumn{1}{l}{H@1} & \multicolumn{1}{l}{H@3} & \multicolumn{1}{l}{H@10} \\
			\hline
%			ComplEx\footnotesize{\cite{Trouillon:2016}} & 0.394 & 30.3  & 43.4    & 57.2   & 0.247  & 15.8  & 27.5  & 42.8 \\
			ConvE\footnotesize{\cite{Dettmers:2018}}    & 0.211  & 13.2  & 23.1 & 36.8    & 0.416 & 32.3  & 45.7    & 60.1 \\
			SACN\footnotesize{\cite{Shang:2019}}     & 0.241  & 15.8  & 26.0 & 40.9   & 0.446 & 35.2  & 49.0    & { 63.1}\\

			RotatE\footnotesize{\cite{Sun:2019}}    & 0.239 & 14.9  & 26.5    & 42.4    & 0.432  & 32.9  & 47.7 & 63.9 \\
%			GATs\footnotesize{\cite{Nathani:2019}}  & 204   & 0.455 & 38.1  & 48.6    & 59.9  & 160   & 0.482  & 40.7  & 51.4 & 62.8 \\
			
%			HAKE\footnotesize{\cite{Zhang:2020}}  & 240   & 0.255 & 16.4  & 28.0   & 44.3  & 127   & 0.436  & 33.3  & 48.4 & 64.2 \\
			InteractE\footnotesize{\cite{Vashishth:2020a}}     &0.258&17.0&28.3&43.7&0.454&{\bf 35.8}&49.8& 64.4  \\
%			CompGCN\footnotesize{\cite{Vashishth:2020b}}     & 0.259 & {\bf 17.4}  & {28.4}   & 43.6     & {\bf 0.455}  & {\bf 35.9}  & {\bf 50.1} & 64.2 \\				
%			CoKE\footnotesize{\cite{Wang:2020}}   & 509   & 0.269& 17.9& 29.9& 45.0& 387& 0.458& 36.1& 50.2& 64.7  \\
			\hline
			{\bf AcrE (Serial)}  & {0.254} & {16.6} & {27.9} &{43.4}   &{0.451} &{35.3} &{49.7} &{64.2}     \\
			{\bf AcrE (Parallel)}  & {\bf 0.261} & {\bf 17.3} & {\bf 28.6} &{ \bf 44.2}   &{\bf 0.455} &{\bf 35.8} &{\bf 49.9} &{\bf 64.7}     \\
			\hline
	\end{tabular}}%
	
	\caption{Head and Tail Predictions on FB15k-237. All the compared results are the best results we can achieve by running the source codes provided by the original papers. }
	\label{tab:detail1}%
\end{table}

\begin{table}[t]
	\centering
	%	\caption{Add caption}
	\begin{tabular}{llcccccccc}
		\toprule
		\multirow{2}[2]{*}{Model} & \multicolumn{4}{c}{Prediction Head (Hits@10)} & \multicolumn{4}{c}{Prediction Tail (Hits@10)} \\
		\cmidrule{2-9}
		\multicolumn{1}{c}{} & \multicolumn{1}{c}{1-to-1} & \multicolumn{1}{c}{1-to-n} & \multicolumn{1}{c}{n-to-1} & \multicolumn{1}{c}{m-to-n} & \multicolumn{1}{c}{1-to-1} & \multicolumn{1}{c}{1-to-n} & \multicolumn{1}{c}{n-to-1} & \multicolumn{1}{c}{m-to-n} \\
		\midrule
		TransE(Bodes et al., 2013) & 43.7  & 65.7  & 18.2  & 47.2  & 43.7  & 19.7  & 66.7  & 50.0 \\
		TransH(Wang et al., 2014) & 66.8  & 87.6  & 28.7  & 64.5  & 65.5  & 39.8  & 83.3  & 67.2 \\
		TransD(Ji et al.,2015)  & 86.1  & 95.5  & 39.8  & 78.5  & 85.4  & 50.6  & 94.4  & 81.2 \\
		CTransR(Lin et al., 2015a) & 81.5  & 89    & 34.7  & 71.2  & 80.8  & 38.6  & 90.1  & 73.8 \\
		KG2E(He et al., 2015) & 92.3  & 94.6  & 66    & 69.6  & 92.6  & 67.9  & 94.4  & 73.4 \\
		TranSparse(Ji et al., 2016) & 87.5  & 95.9  & 44.4  & 81.2  & 87.6  & 57    & 94.5  & 83.7 \\
		STransE(Nguyen et al., 2016) & 82.8  & 94.2  & 50.4  & 80.1  & 82.4  & 56.9  & 93.4  & 83.1 \\
		TransG( Xiao et al., 2016) & {93.0} & 96    & 62.5  & 86.8  & {92.8} & 68.1  & 94.5  & {88.8} \\
		ComplEx\cite{Trouillon:2016} & {\bf 93.9}  & {\bf 96.9}  & 69.2  & 89.3  & {\bf 93.8}    & 82.3  & 95.2  & 91.0 \\
		Jointly(Xu et al., 2017) & 83.8  & 95.1  & 21.1  & 47.9  & 83    & 30.8  & 94.7  & 53.1 \\
		
		RotatE\cite{Sun:2019} & 92.9  & 96.7  & 60.2  & 89.3  & 92.3    & 71.3  & {\bf 96.1}  & {\bf 92.2} \\
		\midrule
		{\bf AcrE(Serial)} & {91.0}  & {\bf 96.9}  & {\bf 70.5}  & {89.1}  & {90.7}    & {\bf 85.2}  & {95.9}  & {\bf 92.2} \\
		{\bf AcrE(Parallel)} & {92.5}  & {\bf 97.0}  & {\bf 69.5}  & {\bf 89.5}  & {92.1}    & {\bf 84.0}  & {\bf 96.1}  & {\bf 92.7} \\
		
		\hline
	\end{tabular}%
	\caption{Predictions by Categories on FB15k. The compared results are taken from their original papers.}
	\label{tab:detail2}%
\end{table}%

\begin{table}[t]
	\centering
	
	%\begin{tabular}{llllll}
	\begin{tabular}{llllllllll}
		\toprule
		\multicolumn{2}{r}{} & \multicolumn{4}{c}{FB15K}  & \multicolumn{4}{c}{FB15k-237} \\
		\cmidrule{2-10}
		\multicolumn{2}{r}{} &  \multicolumn{1}{l}{MRR} & \multicolumn{1}{l}{H@1} & \multicolumn{1}{l}{H@3} & \multicolumn{1}{l}{H@10} & \multicolumn{1}{l}{MRR} & \multicolumn{1}{l}{H@1} & \multicolumn{1}{l}{H@3} & \multicolumn{1}{l}{H@10} \\
		
%		\multicolumn{2}{r}{\multirow{2}[2]{*}{}} & \multicolumn{4}{c}{FB15k/FB15k-237} \\
%		\multicolumn{2}{r}{} & Hits@10 & Hits@3 & Hits@1 & \ MRR \\
%		\multicolumn{4}{c}{FB15k/FB15k-237} \\
%		\multicolumn{2}{r}{} & Hits@10 & Hits@3 & Hits@1 & \ MRR \\
		\midrule
		\multicolumn{2}{l}{\textbf{AcrE(Serial)}} & {0.791} & {72.7} & {83.8} &{89.6} & {0.352}  & {26.0}  & {38.8}  & {53.7}  \\
		\multicolumn{2}{l}{-Residual} & 0.776 & 70.6 & 82.8 & 89.1 & 0.351& 25.8 & 38.6 & 53.7 \\
		\hline
		\multicolumn{2}{l}{\textbf{AcrE(Parallel)}} & {0.815} & {76.4} & {85.2} &{89.8} &{0.358}  & {26.6}  & {39.3}  & {54.5} \\
		\multicolumn{2}{l}{-Residual} & 0.804 & 74.6 & 84.9 & 89.7 & 0.355& 26.1 & 39.0 & 54.1 \\
		\hline

		\multirow{2}[0]{*}{\textbf{AcrE (Parallel)}}   & {\textit{add}}& {0.803} & {74.4} & {84.6} &{89.7} &{0.356}  & {26.5}  & {38.9}  & {54.1}\\
		\multicolumn{1}{l}{} & {\textit{con}} & {0.815} & {76.4} & {85.2} &{89.8} &{0.358}  & {26.6}  & {39.3}  & {54.5}  \\
		
		\hline
	\end{tabular}%
	\caption{Ablation experiments on FB15k and FB15k-237. ``\emph{add}" and ``\emph{con}" refer to the element-add and concatenation integration methods respectively.}
	\label{tab:abl}%
\end{table}%

%\begin{comment}
\begin{table}[t]%[tp]
\centering
%	\small
%	\makebox[\linewidth]
{\begin{tabular}{ll}
\hline
Models &  \multicolumn{1}{l}{ ParaNum (Millions)} \\
\hline
ConvE{\cite{Dettmers:2018}}& $\approx$ {\bf4.96} \\
RotatE{\cite{Sun:2019}}  & $\approx$ 29.32 \\
SACN{\cite{Shang:2019}}  & $\approx$ 9.63 \\
InteractE{\cite{Vashishth:2020a}} &$\approx$ 10.7\\
CompGCN{\cite{Vashishth:2020b}}  &$\approx$ 9.45 \\
HAKE{\cite{Zhang:2020}}   &$\approx$  29.79 \\
CoKE{\cite{Wang:2020}}   &$\approx$ 10.19 \\
%			GATs\footnotesize{\cite{Nathani:2019}}  & 7.62 (4.65/2.97)  \\

\hline
{\bf AcrE (Serial)} &$\approx$ 5.61   \\
{\bf AcrE(Parallel)} &$\approx$ 6.22   \\
\hline
\end{tabular}}
\caption{Parameter efficiency on FB15k-237 (``ParaNum" refers to the number of parameters). }
\label{tab:paras}%
\end{table}% 
%\end{comment}

%Besides, there is another inherent advantage in the \emph{Parallel} structure: it is far faster than the \emph{Serial} structure for both training and inference due to the sequential nature of the \emph{Serial} model. 

\noindent\textbf{Ablation Results} Table \ref{tab:abl} shows the ablation experiments of our model on FB15k and FB15k-237. We can see that there is a large different between the performance of “with/without” residual learning in most cases. As analyzed above, the more serial convolutions used, the more original information would be \emph{forgotten}. While a residual learning adds the original information back. Accordingly, the mentioned issue is  alleviated   greatly. Since \emph{AcrE (Serial)} \emph{forgets} more original information than \emph{AcrE (Parallel)}, it achieves more performance gains from residual learning.

From Table \ref{tab:abl} we can also observe that the integration method  plays important role in  \emph{AcrE (Parallel)}. Usually, the concatenation based integration method is superior to an element-add based integration method in most cases. Here we do not use some complexer integration methods like \emph{gate control} based methods for we do not want to make the model too complex. 

Besides, the atrous rate and the number of atrous convolutions used also affect the performance. Here we do not report the performance under different settings of these two hyper-parameters due to  space limitation. In fact, both of these two parameters are easily selected due to their small search spaces.

\noindent {\bf Parameter efficiency} We also compare the parameter efficiency  between our model and some state-of-the-art models on FB15k-237. For each method, we report the number of parameters associated with the optimal configuration that  leads to the performance shown in Table \ref{tab:main}. The comparision results are shown in Table \ref{tab:paras}, from which we can see that the number of parameters in {\em AcrE} is   close  with {\em ConvE}, but is far less than that in  other compared baselines. This is in line with our expectation: using atrous convolutions would not increase the parameters greatly. These results show that  our model  is more parameter efficient, it achieves substantially better results with fewer parameters. Note that  \emph{AcrE (Parallel)}  has more parameters than  \emph{AcrE (Serial)} because it has an extra transformation operation after the result integration.   %much faster in real prediction than the compared baselines.  %During  testing (or the second phase), the parameters of both models are far less than all the compared baselines. This means that our model (including \emph{GATs}) is faster in real prediction than the compared baselines.

Here we do not quantitatively compare the runtime of different models for it is  difficult to provide a fair evaluation environment: coding tricks, hyper-parameter settings (like  \emph{batch-size}, \emph{learning rate}), parallelization, lot of non-model factors  affect the runtime. However, {\em AcrE} can be viewed as a variant of \emph{ConvE}. Theoretically, it has the same time complexity as \emph{ConvE} that has been proven to be faster than most existing state-of-the-art methods.  Taking FB15k-237 as an example, when using a Titan XP GPU server, it takes about 220 and 100 seconds per epoch during training for \emph{AcrE (Serial)} and \emph{AcrE ({Parallel})} respectively. As for inference, it only takes 14 and 6 seconds  for \emph{AcrE (Serial)} and \emph{AcrE ({Parallel})} respectively to finish the whole test set evaluation.  While some latest GNN or DNN based methods often takes many hours even several days to complete the same work under the same experiment settings. %Both the training and testing  

\section{Conclusions}
In this paper, we propose \emph{AcrE}, a simple but effective DNN-based KGE model. We make comprehensive comparisons between \emph{AcrE} and many state-of-the-art baselines on sis diverse benchmark datasets. Extensive experimental results show that \emph{AcrE} is very effective and it achieves better results than the compared baselines under most evaluation metrics on six benchmark datasets.  The main contributions of our method are summarized as follows. 
First, to our best knowledge, this is the first work that uses different kinds of convolutions for the KGE task. Second, we propose two simple but effective learning structures to integrate different kinds of convolutions together. Third, the proposed model  has much better parameter efficiency than the compared baselines.% and it achieves much better results with fewer parameters. %Besides, our model is very fast and can be easily used in the on-line or real-time application scenarios.  

\section*{Acknowledgements}
This work is supported by the National Key R\&D Program of China (No.2018YFC0830701), the National Natural Science Foundation of China (No.61572120), the Fundamental Research Funds for the Central Universities (No.N181602013 and No.N171602003), Ten Thousand Talent Program (No.ZX20200035), and Liaoning Distinguished Professor (No.XLYC1902057).

\bibliographystyle{coling}
\bibliography{coling2020-AcrE}

\end{document}